\documentclass[letterpaper, 10 pt, conference]{ieeeconf}  

\IEEEoverridecommandlockouts                              

\usepackage{cite}
\usepackage{algorithmic}
\let\labelindent\relax
\usepackage{enumitem}
\usepackage{graphicx}
\usepackage{textcomp}
\usepackage{amsmath,amsfonts,amssymb,amscd}
\usepackage{xcolor}

\title{\LARGE \bf
Reinforcement Learning based dynamic weighing of Ensemble Models for Time Series Forecasting}
\author{Satheesh K. Perepu$^{1}$, Bala Shyamala Balaji$^{2}$,  Hemanth Kumar Tanneru$^{3}$, Sudhakar Kathari$^{4}$, Vivek Shankar Pinnamaraju$^{5}$  
\thanks{$^{1}$ Satheesh K Perepu is with Ericsson Global Services, Chennai, Tamil Nadu, India
        {\tt\small perepu.satheesh.kumar@ericsson.com}}%
\thanks{$^{2}$ Bala Shyamala Balaji is with the Department of Chemical Engineering, Indian Institute of Technology - Madras, Chennai, Tamil Nadu, India
{\tt\small balashyamala@gmail.com}}%
\thanks{$^{3}$ Hemanth Kumar Tanneru is with the Department of Chemical Engineering, Indian Institute of Petroleum and Energy, Visakhapatnam, Andhra Pradesh, India
{\tt\small hemanth.che@iipe.ac.in}}
\thanks{$^{4}$ Sudhakar Kathari is with Honeywell Connected Enterprise, Bangalore, Karnataka, India
{\tt\small sukathari@gmail.com}}
\thanks{$^{5}$ Vivek Shankar Pinnamaraju is an Assistant Professor at Indian Institute of Technology- Dhanbad, Jharkand, India
{\tt\small viveksp@iitism.ac.in}}
}
\begin{document}

\maketitle
\thispagestyle{empty}
\pagestyle{empty}

\begin{abstract}
Ensemble models are powerful model building tools that are developed with a focus to improve the accuracy of model predictions. They find applications in time series forecasting in varied scenarios including but not limited to process industries, health care, and economics where a single model might not provide optimal performance. It is known that if models selected for data modelling are distinct (linear/non-linear, static/dynamic) and independent (minimally correlated models), the accuracy of the predictions is improved.  Various approaches suggested in the literature to weigh the ensemble models use a static set of weights. Due to this limitation, approaches using a static set of weights for weighing ensemble models cannot capture the dynamic changes or local features of the data effectively. To address this issue, a Reinforcement Learning (RL) approach to dynamically assign and update weights of each of the models at different time instants depending on the nature of data and the individual model predictions is proposed in this work. The RL method implemented online, essentially learns to update the weights and reduce the errors as the time progresses. Simulation studies on time series data showed that the dynamic weighted approach using RL learns the weight better than existing approaches. The accuracy of the proposed method is compared with an existing approach of online Neural Network tuning quantitatively through normalized mean square error(NMSE) values.

\end{abstract}

\section{Introduction}
\label{sec:intro}

Model building is a fundamental step in understanding a process, estimating the optimal parameters, making predictions, fault detection, safety and economic aspects of any process.  There are 2 primary approaches to modelling (i) first-principles models, where the mathematical model of the process is derived from fundamental physical laws, and (ii) data-driven models, where the model that describes the dynamics of underlying process is estimated from the data. In this paper, we focus on the latter approach for time series forecasting. Although there exists different data-driven methods and model structures \cite{book:ljung,kim2019time,wang2018novel,jiang2017design}, forecasting industrial data using a single model structure may not be sufficient to achieve the desired performance which brings out the two questions as below. 
\begin{enumerate}
    \item Which is the best model amongst all models built for the given data? \label{it:A}
    \item Will all the models built in \ref{it:A} provide a similar performance on entire data set?
\end{enumerate}
To answer the first question, generally a trial and error approach is employed. Different sets of models are built and are validated using predictions on the test data. Model with least prediction error is naturally selected as the best one. However, it doesn't mean that the chosen model is able to capture the data characteristics well. If the data contains different local features such as linear/periodic etc., different types of models are required to capture them. However, identifying all the features in the data is strenuous owing to complex nature of data. Hence it is difficult to compare models built in (A), which answers the second question. In order to improve the performance of the resulting model, recently, building an ensemble of individual models has gained attention. However, the accuracy of an ensemble model depends on the weights given to the individual models since the contribution of models may vary based on the local features of the data. Hence, it is important to decide how to select weights for the different models under consideration. There are various ways in which the ensemble models are generated. 

Ensemble models in the literature are  predominantly built based on two approaches: i) serial and ii) parallel. The serial approach involves dividing the training data into subsets and building different models for each split. For instance,  Zhang et.al. \cite{paper:time_Series_Serial_exchange_data} has provided an approach for predicting exchange rates using a serial ensemble of neural network models. To use this method, some primary knowledge about local features and data trend is essential to split into different regions for model building. On the other hand, the parallel approach ensemble models are built on the entire training data\cite{paper:ensemble}, i.e., many relevant models are built for the entire set of data. In both the approaches, the predictions obtained from individual models are combined to arrive at a single prediction. 

A straightforward ensemble would be to use the first modal value or a simple average of predictions of all models for a regression problem or a majority vote for a classification problem. An alternate approach is to compute a weighted average of individual models by solving an optimization problem to minimize the overall prediction error for a regression problem or to minimize the misclassification rate in the case of a classification problem. The disadvantage of these approaches is that they use constant value of  weights through out the entire training data for the different models under consideration. However, this may not be correct since the models' performance depends on local features of the data. Based on the individual model performance, the weights of the ensemble models are required to be updated when a new sample arrives. To address the same, in this work we resort to using  Reinforcement Learning (RL) to dynamically estimate the weights of the ensemble model.

RL derive their foundations from the aspects of optimal control. The basic aim of RL is to maximize the reward (which is measure of how good your system performs) obtained by performing the actions (input in control terminology) such that the system moves from one state to another state (direction that yields good reward). For more details on RL, reader is directed to a book by Sutton and Barto \cite{book:sutton2018reinforcement}. Few works in literature have also reported the usage of  RL techniques for pruning of ensemble models \cite{book:RL_ensemble,thesis:RL_ensemble2}. In these works, RL techniques are used for ensemble pruning i.e. to select which model is performing better for the current sample in the case of classification problem. The action space (weights of individual model) and state space (prediction error) as considered as discrete, specifically binary. However, for the case of forecasting, the state space and action space should be infinite since the error and weights can take any real value. For the ease of solving the problem, we assume the action space is infinite and state space is finite. More details on this is explained in the Section \ref{sec:prob_methodology}.

Addressing the challenges as above, the novel contributions of the work are i) to pose the problem of time series forecasting as weighted ensemble model, ii)to compute weights of the ensemble model dynamically by using RL technique assuming infinite action space. It is to be kept in mind that RL is not directly used for the model predictions. The  advantage of this approach is that RL can dynamically learn the different model weights depending on the local features of the training data at each instant of time. This can help us in time series forecasting or missing data identification after sufficient training using RL. The method at each instant, rewards or penalizes each model depending on the prediction at previous instant(s), which in turn is an aggregate of the past prediction errors.

The remainder of the  paper is organized as follows. Section \ref{sec:prob_methodology} focuses on explaining the proposed approach using the RL methodology and a Neural Network based dynamic weighting of ensemble models with which the results are compared. Section \ref{sec:results} provides the results on a time series data set and discussion on the performance of the proposed approach. This is followed by Section \ref{sec:conclusions} providing the contributions of the proposed approach and the possible extensions to future work.

\section{Problem formulation}
\label{sec:prob_methodology}

A general description of ensemble model prediction is explained below. Let $\mathbf{y}$ be an univariate time series data in $\mathbb{R}^n$ and the prediction of the time series data at an instant of time $t$ using a model $i$ is given by $\hat y[t]^i$. The predictions from different models is combined linearly as in Eq. \eqref{eq:ensemble}.
\begin{align}
     {\hat y}_p[t] = {\sum_{i=1}^M w_i {\hat y}^i[t]}
    \label{eq:ensemble}
\end{align} where  $\mathbf{y}\in \mathbb{R}^N$ is uni-variate time series data, $M$ is the number of models, $\hat {y}^i [t]$ represents the prediction from the $i^{th}$ model at time instant $t$ and $w_i$ represents the weights used to combine predictions. The individual predictions are weighted averaged out to obtain the combined prediction $\hat y_p[t]$ as per Eq. \eqref{eq:ensemble}. The individual weights are chosen to be between $[0,1]$ and the sum  of the weights amounting to $1$. One can solve this as optimization problem and we compute the static set of weights for entire training data. However, the formulation has the disadvantage that the individual models can perform differently across the entire stretch of training data.


Hence it is better to dynamically identify the weights of the model depending on the local features of the data. The RL method can learn new data trends to effectively modify weights to improve the forecast. Having described the problem statement, we will focus on the interpretation of the RL terminologies with respect to the problem under consideration and the challenges and solution methods. 

The important terminologies in RL are the reward, policy, states and actions which are explained as follows. 
\begin{enumerate}
    \item \textit{States} : States describe the output of the system i.e. the system response. Here states are considered as the normalized form of the combined prediction error of the ensemble of models expressed as percentage. If we consider the number of models are $M$ for training the system. Then the current state is given by
    \begin{align}
        S_t = \frac{ (y[t]- \displaystyle\sum_{i=1}^M w^i[t]\hat{y}^i[t])^2}{y[t]^2} \times 100
        \label{eq:states}
    \end{align}
    where $y[t]$ is the true value of prediction at time $t$ and $\hat{y}[t]^i$ is the prediction of the $i_{th}$ model at time $t$ weighted by the factor $w_i$ at time $t$. 
    
    As already mentioned in Section \ref{sec:intro}, the computational cost of the RL methods depends on number of states and actions of the process. Here, the prediction error is considered as state of the process and can take infinite values. Hence, we choose to discretize the prediction error into $n$ categories to make the states finite. The finite state given as $S_{ft}=LB$ when $LB \leq S_t \leq UB$ and the $LB$ and $UB$ are the obtained by dividing [0,100] to $n$ intervals.
The process is considered as a Markov Decision process wherein the next state depends only on the state action pair at the current instant of time. 
    \item \textit{Actions} : Actions represent the movement  from one state to another. With respect to this work actions are the weights assigned to each of the $M$ models. The actions are considered to be an infinite variable in this approach. Although, this results in high computational cost, there is no way we can discretize them as a small change in weight can lead to large changes in error of the predictions. Hence, the actions/weights to each of the ensemble models  is a continuous value ranging from $0$ to $1$. Weight $w^i[t]$ in Eq. \eqref{eq:states} is denoted as action $A_t$ for the RL problem. 

To summarise, the prediction error forms the state and the assignment of weights become the actions. Next we will look into the concepts of the reward function which forms the objective in the RL framework. 

    \item \textit{Reward and Return} : Return is the total discounted reward obtained due to the current action performed to move the system output from current to the next state. The objective of the work is to apply actions such that the system is moved to a state which yields maximum return. In RL, the return at the current instant depends not only on the instantaneous reward obtained but also on the future rewards.  This can be described as 
    \begin{align}
        G_t = R_{t+1}+\gamma R_{t+2}+\gamma^2 R_{t+3}\cdots = \sum_{k=1}^{\infty}\gamma^k R_{t+k+1} 
    \end{align}
where $R_t$ is the reward at current time $t$ and $\gamma$ is a scalar between $[0,1]$. $\gamma$ is known as the discounting factor and is used to give importance to the contribution of reward to the future states due to the current action. In other words, if $\gamma$ is close to $1$, we give more importance to the future states and if the $\gamma$ is close to $0$, lower importance is given to the farther states. For the proposed method, the reward at an instant $t$ is computed as 
\begin{align}
    R_t = S_{t-1}-S_t 
    \label{eq:reward}
\end{align} where $S_t$ is representative of the prediction error at instant $t$. Naturally, it is required to lower the prediction error and hence this reward function is put to use. More the decrease in the prediction error, better is the reward. 
Different reward forms like inverse of prediction error ($1/|S_t|$), inverse of difference of prediction error ($1/|S_t-S_{t-1}|$) can also be used. However, from simulations, Eq. \eqref{eq:reward} was found to perform better.   
\item \textit{Policy} : Policy provides the distribution of actions for the current state. A policy $\pi$ is given by
\begin{align}
    \pi (a|s) = \mathbb{P}[A_t = a|S_t = s].
\end{align}
It is required to identify the optimal policy to maximize the rewards. Using the distribution of actions we can come up with state and action value functions which give us the long term value of states and expected reward to move from current state and action following the policy function. 
The state value function is given by
\begin{align}
    v_\pi(s) = \mathbb{E}_\pi[G_t|S_t = s].
\end{align} and the action value function is given by
\begin{align}
    q_\pi(s, a) = \mathbb{E}_\pi[G_t|S_t = s, A_t = a].
\end{align}
The optimal state/action value function is given as the maximum of the state/action value function. The objective of the RL is to find optimal policy which generates total maximum reward. 

There are two types of policy learning methods in literature \cite{book:sutton2018reinforcement} (i) off-policy learning, like Q-learning where the value function is learned from executing the actions of another policy and (ii) on-policy learning, like SARSA, where the value function is learned from executing actions of same policy. For detailed analysis of these methods, readers are advised to read \cite{book:sutton2018reinforcement}. In this work, we used online policy learning to update the policy over the episodes.
\end{enumerate}

Delving into to the major challenges in using RL, in certain cases it is not trivial to arrive at reward function as the system can be a black box model. In such cases, we resort to deep-RL where the reward function is modeled using a deep neural network. The deep network is trained with states as input and actions and rewards as the output. The deep neural network solves a regression problem to estimate best action which generates high reward for a given state. The regression problem ensures mapping the infinite action space which translates to infinite choice of weights. The weights of the network are updated based on the predicted value function from the Bellman equation and actual reward obtained from the network. In this way, the proposed method helps in identifying the weights of the proposed method.
 
\begin{figure*}
    \centering
    \includegraphics[scale=0.6]{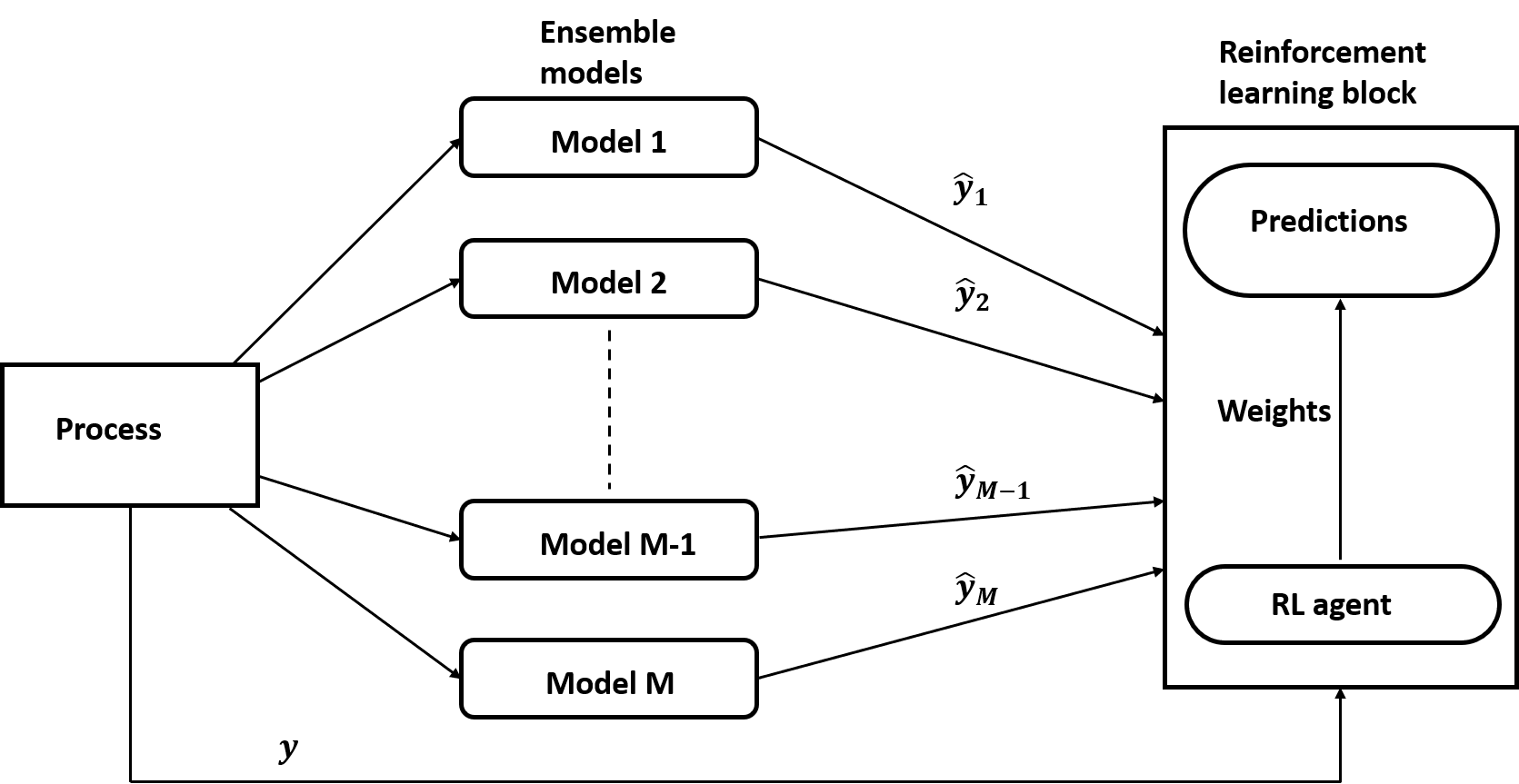}
    \caption{Proposed methodology using Reinforcement Learning}
    \label{fig:method}
\end{figure*}

The methodology of the weight updation using RL can be explained using Fig. \ref{fig:method}.
The following steps are performed in order to compute the predictions using the proposed approach 
\begin{enumerate}
    \item $\mathbf{y} \in \mathbb{R}^N$ is divided into $\mathbf{y}_{train} \in \mathbb{R}^K$ to train the ensemble models and RL and $\mathbf{y}_{test} \in \mathbb{R}^{N-K}$ to test the predictions $\mathbf{y}_p$   
    \item Fit  $M$ different models for $\mathbf{y}_{train}$ and estimate the predictions $\mathbf{\hat y}^i ~\forall ~i \in {1,2,\cdots,M}$ for $N-K$ samples. 
    \item Formulate the RL problem and use episodic RL learning with $P$ episodes by dividing $\mathbf{y}_{train}$ as $[\mathbf{y}_{train,1} \mathbf{y}_{train,2} \cdots \mathbf{y}_{train,P}]^T$ where $\mathbf{y}_{train,i} \in \mathbb{R}^{K_b}$ and $K_b = \frac{N}{P}, K_b \in \mathbb{Z}$.  Selection of $K_b$ depends on the size of training data. 
    \item The output of the RL model will be a dynamic set of weights for the ensemble models on each instant of training data in each episode. Reset the system after each episode.
    \item Use the learned RL model to predict the instant by instant of the first sample of the testing data and compute the prediction error.
    \item Input the error to the RL model and obtain the new set of weights which is used to predict next instant of the test data. 
    \item Repeat steps 5-6 until all the test data is predicted.
\end{enumerate}

 An interpretation of the proposed method can be seen as online model which updates weights of ensemble models at  every instant of time. The weights of the ensemble models are updated not only based on the previous instant prediction error but also on the instants before that. The weights of ensemble models can be seen as non-linear function of the prediction errors of previous instants and function is continuously updated at every instant. For demonstrating the efficacy of the method, we compare the results of proposed method with another famous non-linear approximation (neural network (NN)) to update the weights of ensemble models. More details on the results is discussed in next section of the paper. 
 

\section{Results and Discussion}
\label{sec:results}

The proposed approach has been tested on the benchmark CATS (Competition on Artificial Time series) dataset \cite{paper:CATS} which was released as part of IJCN 2004 Time series prediction problem. The total length of the time series data  is $5000$ samples out of which $100$ data points are considered missing and these missing samples are grouped into 5 blocks as $981$ to $1000$, $1981$ to $2000$, $2981$ to $3000$, $3981$ to $4000$ and $4981$ to $5000$.

The goal of this problem is to predict those $20 \times 5$ missing data points. Since, we are trying to solve this problem using time series forecasting approach, we predict those samples only using the previous samples. For the sake of comparison, we also predict the missing values in the dataset using the online learning NN method where the weights are updated at every sample since the proposed method with RL also updates the weights at every sample. In general, this may not be efficient as updating the weights for a batch of samples which can lower the computational level. This is because, the features of data may not change from sample to sample. However, this is beyond the scope of this work and can be solved by adding batch size as one of the hyperparameters to the problem.

The ensemble of models selected for this data are a linear regression model, Long Short Term Memory (LSTM) model, Artificial Neural Networks (ANN) and Random forest. The different models selected ensure that it can capture linear trends if the data is linear (linear regression) and to capture lower order non linearity (ANN with 2 hidden layers), LSTM model to take the past data into effect and the random forest as it can give a good performance with least interpretation. As discussed, $20$ data points are missing for every $1000$ samples and hence we consider every $980$  samples as training data and subsequent $20$  samples as test data. These $980$ samples are considered as individual samples and which means we have $980 \times 5$ samples as training data and intermediate missing  $20\times 5$ form test data.  For each $980 \times 1$ samples, we fit $4$ different models as mentioned above. Similarly, we fit for every set of training data to obtain five sets of 4 different types of models. The parameters of the models for different sets of data are averaged to obtain the $4$ different types of models representing the entire training data. 

\begin{figure}
    \centering
    \includegraphics[width=8cm,height=6cm]{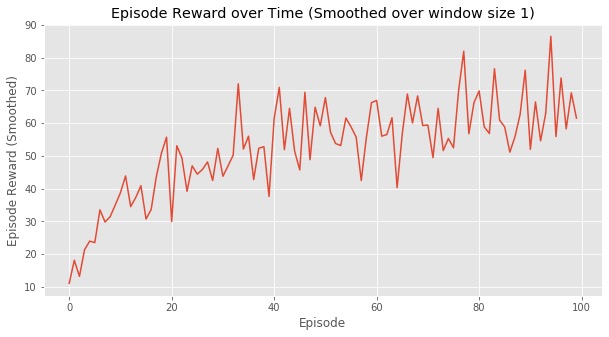}
    \caption{Smoothed reward function over different episodes}
    \label{fig:episodes}
\end{figure}

 Coming to the computation details of the problem, each of the $980$ samples are considered as single episode in RL and total training samples ($980 \times 5$) are considered as $5$ episodes. We run the RL for $100$ episodes which is equivalent to running through the entire training data $20$ times. The reward obtained across each episode is summed up and total reward is plotted with respect to episodes in Fig. \ref{fig:episodes}. It can be seen from the Fig. \ref{fig:episodes} that as the number of episodes increase, the reward value obtained also tends to increase. This is because the system is learning to  modify the coefficients (weights) such that the prediction error decreases with every episode. 
 
\begin{figure}
    \centering
    \includegraphics[scale=0.6]{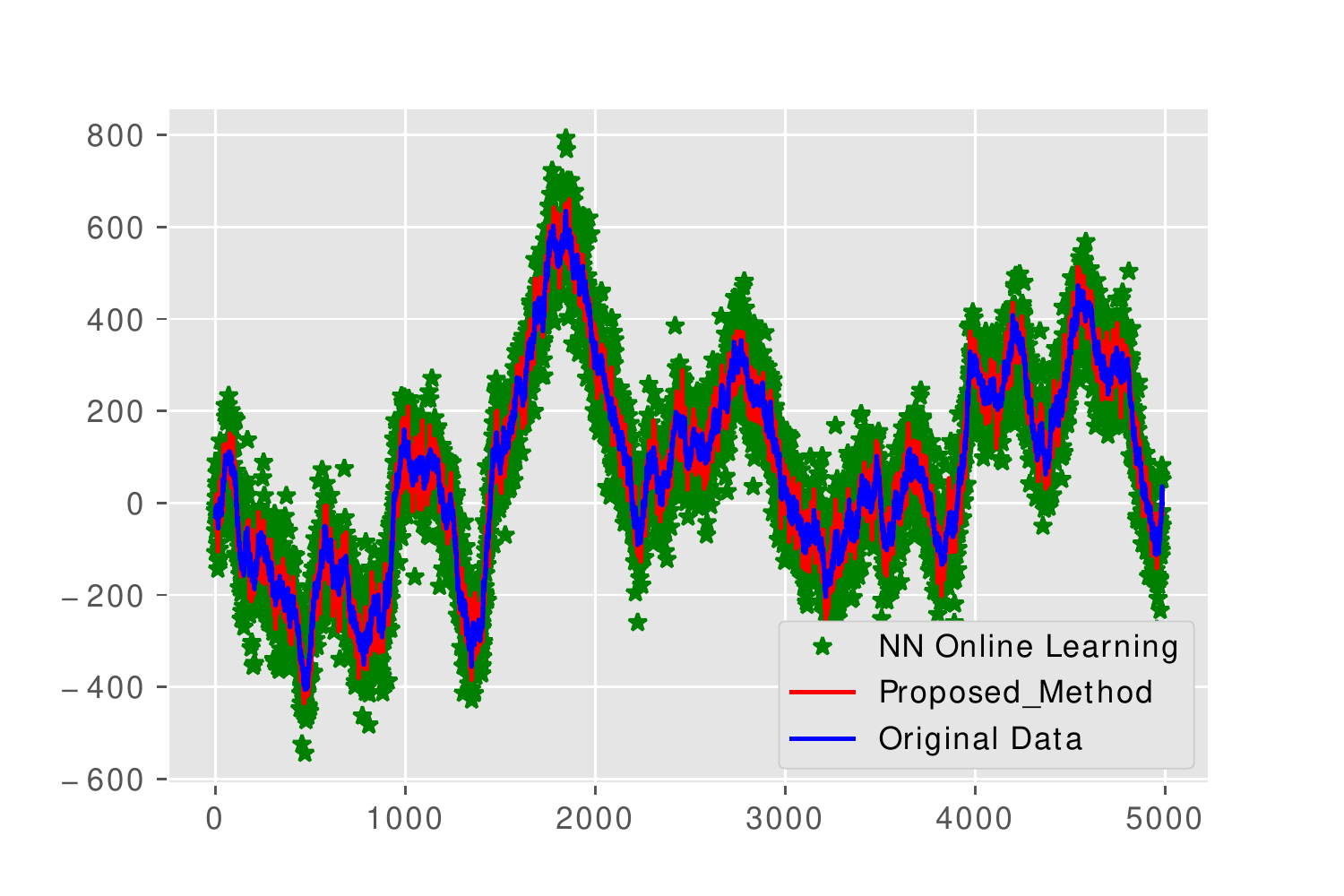}
    \caption{Comparison of predictions of true data to RL based proposed approach and Online NN}
    \label{fig:proposed_method}
\end{figure}

Though the purpose of the work is not to use different models for different regions of data, it was observed that different models performed better in different regions of data. The performance is decided by observing the sum of weights given to the various models in different bands (considered here as $1000$ samples) of data. We have considered the model which has the maximum weightage in each band as the dominant model. It was observed that in the samples $0-1000$ the Neural Network model worked better whereas in the region $1001-2000$ the Linear model worked better. Similarly in the region $2001-3000$, the LSTM model was better and in $ 3001-4000$, the Neural Network model was again dominant. Finally, in the region $4001-5000$ the Random Forest performed well. All these model identification are not visually evident from the data but a few can be inferred. For instance, it can be observed from Fig. \ref{fig:proposed_method} that in the region around $1400-2000$ there seems to be a linear trend and hence the Linear model performing well in that region is anticipated. 

As explained at the beginning of the section, the weights for the purpose of comparison has been obtained by using online NN. To explain, the same set of $4$ ensemble models are built for each of the $980$ samples and the corresponding model coefficients are averaged to come up with a single model for each ensemble models. Then, we formulate the NN with two hidden layers with 4 nodes each with \textbf{tanh} activation function and 4 input nodes corresponding to the number of ensemble models and 4 output nodes with weights of the models. Initially, the network is initialized to random weights and use the stochastic gradient descent to update the weights of the model. The loss function here takes the estimated output (weights of the ensemble model), computes the prediction at that instant and compares with the true data at that instant. The updation happens at every sample and with every sample we obtain new set of weights which is used to predict the next sample. In the case of the test data, the final updated set of weights are used to predict each sample in contrast to the RL technique where the set of weights are updated even when progressing through the test data. The reason is becasue of the udpated nature of the RL method.  

\begin{figure}
     \centering
     \includegraphics[scale=0.6]{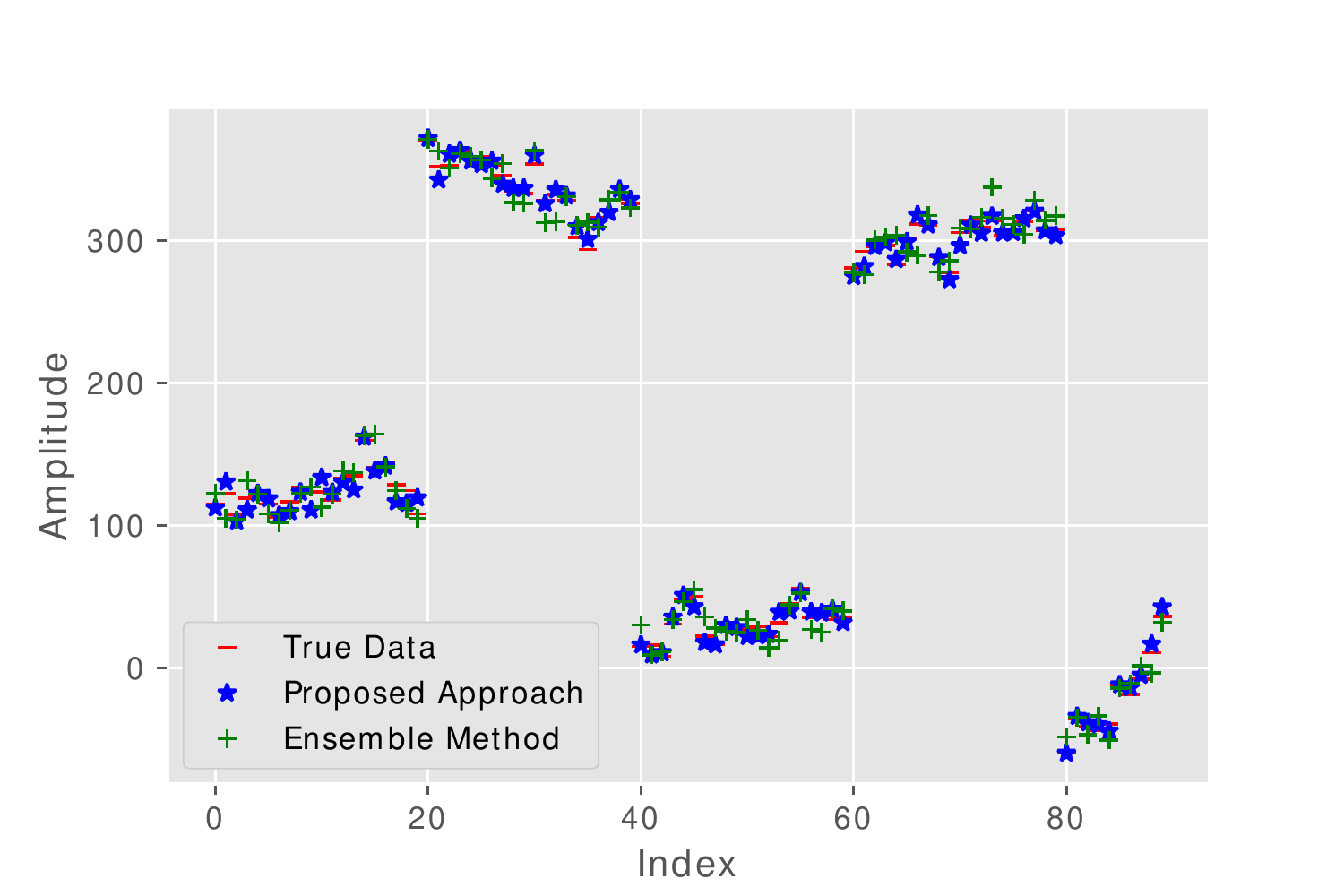}
     \caption{Comparison on the test data}
     \label{fig:test}
 \end{figure}

Figure \ref{fig:proposed_method} compares the true value of time series data with that of predictions obtained from the proposed methodology and online NN.
From the plot, it is evident that the proposed methodology gives predictions closer to the true data when compared with online NN approach. To quantify the performance, we used the NMSE metric to compare both the methods. The metric is computed as
\begin{align}
    \text{NMSE} = \dfrac{\displaystyle\sum_{t=1}^{n_{\text{test}}}(y_t-\hat{y}_{t})^2}{\displaystyle\sum_{i=1}^{n_{\text{test}}}y_t^2}
\end{align}
where $n_{\text{test}}$ is the number of test data samples, and $\hat{y}$ and $y$ are the predicted and the true value of the time series data at time instant $t$. To check the usability of these models on the testing data, we computed NMSE only for the testing samples. Table \ref{table:NMSE} provides the NMSE scores obtained from the two methods and the  individual model performances on the test data i.e., on $100$ samples. From the table it is evident that the proposed ensemble approach with RL outperforms individual models as well as online NN method in terms of the NMSE scores. 
\begin{table}[]
    \caption{NMSE scores for different models}
    \label{table:NMSE}
    \centering
    \begin{tabular}{|c|c|}
    \hline
    \textbf{Model}     &  \textbf{NMSE scores} \\
    \hline
    LSTM     &  0.478\\
    ANN &  0.26 \\
    Linear regression & 0.75\\
    Random Forest &  0.84\\
    Online NN & 0.256\\
    Proposed approach &  0.143\\ \hline
    \end{tabular}
\end{table}
It can be observed from Table \ref{table:NMSE} that ANN model has the maximal accuracy among individual models. The proposed method has a better NMSE of $0.14$ which could be tuned to obtain better scores by modifying the hyperparameters of the RL model. For a clear picture on the performance of the two methods on the test data, Fig. \ref{fig:test} has been included. It is to be observed that the index refers to the indices of the missing variables amounting to $100$ samples.

The computational complexity of the proposed method is equivalent to that of online NN counterpart. In this case study, it took an average of $0.2$ seconds to compute the weights using the RL based technique and $0.14$ seconds to estimate weights for online NN method. These results are obtained when the code is run on local machine with MAC OS with 16GB RAM and i9 processor with no GPU. From the computational time taken, it can be seen that the proposed method can be employed in online applications even with lesser sampling time i.e. less than that of 0.2 seconds. 
From the results, it can be concluded that the proposed method with RL gives good and accurate predictions when compared with the online counterparts. 

\section{Conclusions}
\label{sec:conclusions}

In this paper, a method based on RL is proposed to estimate the weights of an ensemble model dynamically for time series forecasting. Simulation results showed that the proposed method performs well when compared to the existing methods that use NN based dynamic weighting. Further, the proposed method can handle local features of the time series data since the weights are computed dynamically. An added advantage of the proposed method is that the user will be able to identify the predominant model  at any instant of time based on the computed weights. The NMSE scores for different models in modeling the benchmark CATS data set illustrate that the proposed RL-based approach outperforms the existing static weighting methods and online learning methods. 

Future work includes the extension of the proposed method for modeling and forecasting of multivariate time series. Modeling multivariate time series is not an easy task due to the existence of multicollinearity and cross-correlation among predictors. The possible scope of the improvement to the proposed method in modeling univariate time series is that the number of episodes of RL can be reduced using better reward value, which is also an interest of future work. 


\bibliographystyle{unsrt}
\bibliography{CSS}

\begin{thebibliography}{10}

\bibitem{book:ljung}
Lennart Ljung.
\newblock System identification.
\newblock {\em Wiley Encyclopedia of Electrical and Electronics Engineering},
  pages 1--19, 1999.

\bibitem{kim2019time}
Minju Kim, Kosuke Nishi, Kandhasamy Sowndhararajan, and Songmun Kim.
\newblock A time series analysis to investigate the effect of inhalation of
  aldehyde c10 on the human eeg activity.
\newblock {\em European Journal of Integrative Medicine}, 25:20--27, 2019.

\bibitem{wang2018novel}
Yongjian Wang and Hongguang Li.
\newblock A novel intelligent modeling framework integrating convolutional
  neural network with an adaptive time-series window and its application to
  industrial process operational optimization.
\newblock {\em Chemometrics and Intelligent Laboratory Systems}, 179:64--72,
  2018.

\bibitem{jiang2017design}
Dazhi Jiang, Jian Gong, and Akhil Garg.
\newblock Design of early warning model based on time series data for
  production safety.
\newblock {\em Measurement}, 101:62--71, 2017.

\bibitem{paper:time_Series_Serial_exchange_data}
G~Peter Zhang and VL~Berardi.
\newblock Time series forecasting with neural network ensembles: an application
  for exchange rate prediction.
\newblock {\em Journal of the operational research society}, 52(6):652--664,
  2001.

\bibitem{paper:ensemble}
Daijin Kim and Chulhyun Kim.
\newblock Forecasting time series with genetic fuzzy predictor ensemble.
\newblock {\em IEEE Transactions on Fuzzy systems}, 5(4):523--535, 1997.

\bibitem{book:sutton2018reinforcement}
Richard~S Sutton and Andrew~G Barto.
\newblock {\em Reinforcement learning: An introduction}.
\newblock MIT press, 2018.

\bibitem{book:RL_ensemble}
Ioannis Partalas, Grigorios Tsoumakas, Ioannis Katakis, and Ioannis Vlahavas.
\newblock Ensemble pruning using reinforcement learning.
\newblock In Grigoris Antoniou, George Potamias, Costas Spyropoulos, and
  Dimitris Plexousakis, editors, {\em Advances in Artificial Intelligence},
  pages 301--310, Berlin, Heidelberg, 2006. Springer Berlin Heidelberg.

\bibitem{thesis:RL_ensemble2}
Christos Dimitrakakis.
\newblock Ensembles for sequence learning, 01 2007.

\bibitem{paper:CATS}
Amaury Lendasse, Erkki Oja, Olli Simula, and Michel Verleysen.
\newblock Time series prediction competition: The {CATS} benchmark.
\newblock {\em Neurocomputing}, 70:2325--2329, 08 2007.

\end{thebibliography}
\end{document}